\documentclass[journal]{IEEEtran}

\ifCLASSINFOpdf
\else
   \usepackage[dvips]{graphicx}
\fi
\usepackage{url}

\usepackage[dvipsnames]{xcolor}
\usepackage{graphicx}
\usepackage{multirow}
\usepackage{cite}
\usepackage{subfigure}
\usepackage{parskip}
\usepackage{indentfirst}
\usepackage{amsmath}
\begin{document}

\title{POBEVM: Real-time Video Matting via Progressively Optimize the Target Body and Edge}

\author{Jianming Xian}

\markboth{Journal of \LaTeX\ Class Files, Vol. 14, No. 8, August 2015}
{Shell \MakeLowercase{\textit{et al.}}: Bare Demo of IEEEtran.cls for IEEE Journals}
\maketitle

\begin{abstract}
Deep convolutional neural networks (CNNs) based approaches have achieved great performance in video matting. Many of these methods can produce accurate alpha estimation for the target body but typically yield fuzzy or incorrect target edges. This is usually caused by the following reasons:  1) The current methods always treat the target body and edge indiscriminately; 2) Target body dominates the whole target with only a tiny proportion target edge. For the first problem, we propose a CNN-based module that separately optimizes the matting target body and edge (SOBE). And on this basis, we introduce a real-time, trimap-free video matting method via progressively optimizing the matting target body and edge (POBEVM) that is much lighter than previous approaches and achieves significant improvements in the predicted target edge. For the second problem, we propose an Edge-L1-Loss (ELL) function that enforces our network on the matting target edge. Experiments demonstrate our method outperforms prior trimap-free matting methods on both Distinctions-646 (D646) and VideoMatte240K (VM) dataset, especially in edge optimization. 
\end{abstract}


\IEEEpeerreviewmaketitle

\section{Introduction}

\IEEEPARstart{V}{ideo} matting aims to predict the alpha mattes of each frame in the video, with a strong practicality. Formally, a frame I can be view as a linear combination of a foreground image F and a background image B by an $\alpha $ factor:

\begin{equation}
I = \alpha F + (1 - \alpha)  B
\end{equation}
The focus of the video matting task is to accurately predict the $\alpha_i$ value of each pixel i in each frame of the video, where $\alpha_i $ $\in $ [0, 1]. In recent years, with the rapid development of deep learning, deep convolutional neural networks (CNNS) \cite{1998Gradient} have achieved great performance in video matting, and these methods can be generally divided into two categories: Auxiliary-based matting and Auxiliary-free matting. 

\textit{Auxiliary-based matting}: There are general two kinds of auxiliary-based matting methods: trimap-based matting and background-based matting, . 1) Trimap-based matting methods \cite{2009Environmental, Chen2013KNN,2001A,2021Trimap,2020Context,2022Boosting,2021Improving,2022MatteFormer} use a manual trimap annotation, which explicitly defines the known foreground and background as well as unknown parts that need the matting method to solve, as an auxiliary guidance input to help the model extract $\alpha $ of the image. Xu {\em et al.} \cite{2017Deep} were the first to use deep neural networks for trimap-based matting, and many subsequent studies have continued to follow this approach. But this method is time-consuming and labor-intensive, and the quality of the trimap annotation has a significant impact on the model training results. 2) Background-based matting methods \cite{2020Background,2020Real,2022Adaptive} require an additional pre-captured background image as auxiliary guidance input. This approach is more instructive and can effectively improve the matting performance. The BGMv2 \cite{2020Background} proposed by Lin {\em et al.} is one of the representative works of this approach. But this method is not suitable for dynamic matting because it requires the background image and the original input image must be aligned.

\textit{Auxiliary-free matting}: Mainly refers to the methods that do not require any auxiliary input, typical as trimap-free methods \cite{2021Robust,Li2022VMFormerEV,2020MODNet,2022PP,Sun2022HumanIM,Song2022SGMNetSG}. Trimap-free methods often can not achieve performance as well as the trimap-based matting methods, but have greater utility in the field of matting and also have been used more successfully in practical applications. Chen \emph{et al.} proposed PP-Matting \cite{2022PP} combining high-resolution and low-resolution branches which can achieve comparable results to the trimap-based matting methods.  Lin \emph{et al.} proposed RVM \cite{2021Robust} which added the temporal model to the matting network and can process 4K and HD high-resolution images in real time. Sun \emph{et al.} proposed InstMatt \cite{Sun2022HumanIM} which can predict a precise alpha matte for each human instance in an image. All these methods have achieved significant improvement in the prediction of the target body and are comparable to trimap-based methods in various matting evaluation metrics. However, these methods also usually produce ambiguous or even erroneous edge prediction results. This is usually caused by the following two reasons: 1) The current methods always treat the target body and edge indiscriminately; 2) Target body dominates the whole target with only a tiny proportion target edge. The above two reasons lead the model to focus mainly on the target body, thus ignoring the target edge. 

\indent In this letter, inspired by PFNet \cite{2021Camouflaged}, UIM \cite{2022Unified} and TINet \cite{Zhu_Zhang_Zhang_Liu_2021}, we propose a CNN-based block that separately optimizes the matting target body and edge (SOBE) to solve the above problems and validated its performance on three camouflaged object datasets. Based on this SOBE block, we propose a trimap-free network for real-time video matting by progressively optimizing the matting target body and edges (POBEVM). To the best of our knowledge, POBEVM is the first matting method to focus on optimizing target edges. The major contributions of this letter are summarized as follows.
\begin{itemize}
\item We propose a CNN-based SOBE block for optimizing the matting target body and edge separately, and innovatively design a real-time matting network POBEVM.
\item We propose an Edge-L1-Loss (ELL) function that enforces our network on the matting target edge.
\item Extensive experiments show that the proposed POBEVM network can achieve state-of-the-art performance compared to prior trimap-free matting methods on both Distinctions-646 (D646) and VideoMatte240K (VM) dataset, and the proposed SOBE block is highly effective in refining target bodies and edges.
\end{itemize}

\begin{figure}
\centerline{\includegraphics[width=\columnwidth]{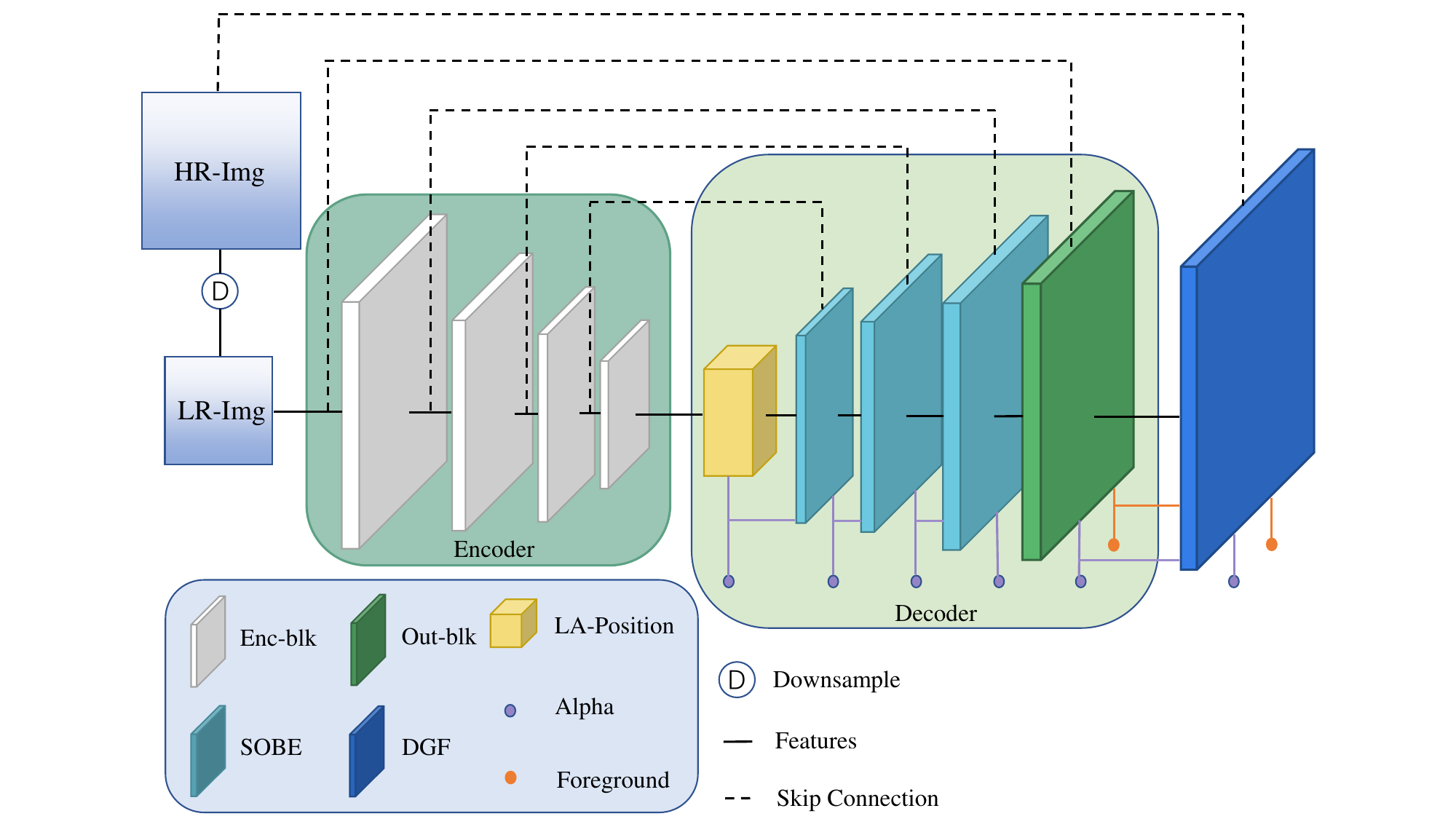}}
\caption{Detailed implementation of the proposed POBEVM network structure. There are five alpha predictions and one foreground prediction. And, when the DGF module is used, its outputs will replace the outputs of the Out-blk.}
\end{figure} 

\section{METHOD}

In this part, we will introduce the structure of our proposed POBEVM network and the training strategy we adopted in detail

\subsection{POBEVM}

The POBEVM network architecture proposed in this study consists of an encoder that extracts image features, a decoder that progressively optimizes the matting target body and edge, and a Deep Guided Filter (DGF) module derive from RVM \cite{2021Robust} for high-resolution upsampling.The whole framework of POBEVM is shown in Fig. 1.

\subsubsection{Encoder}
{To achieve real-time matting, we adopt MobileNetV3-Large \cite{2019Searching} as our encoder which extracts features at $\frac{1}{2}$, $\frac{1}{4}$ , $\frac{1}{8}$, and $\frac{1}{16}$ scales. These extracted features will be fed to the SOBE block in the decoder via a skip connection.}

\subsubsection{Decoder}
{As shown in Fig. 1, our decoder consists of a LA-Position block, SOBE blocks, and a output block(Out-blk).

\emph{LA-Position block} is a combination of the LR-ASPP module \cite{2019Searching} and Position module (PM) \cite{2021Camouflaged}. PM consists of a channel attention block and a spatial attention block aims to generate the initial alpha matte prediction using semantically enhanced high-level features. These two attention blocks are implemented in a non-local way \cite{2021Camouflaged}. Therefore, to reduce the amount of computation, we added an LR-ASPP module for reducing channels before PM. The LR-ASPP module and Position module follow exactly the form in their original paper.

\emph{SOBE block} which takes the current-level features Fc derived from the backbone and the higher-level features Fh and $\alpha$ matte prediction as the input and outputs the refined features and an optimized $\alpha$ matte prediction consists of two branches. We treat the higher-level $\alpha$ matte prediction as a guide map similar to the trimap which allows the model to optimize the edges better. As shown on the left in Fig. 2, after upsampling the higher-level prediction map, multiply it by the current-level features to get the edges-enhanced features Fee. After that, we feed the Fee into a convolution layer which followed by a batch normalization (BN) layer and a ReLU nonlinearity operation (CBR). Then, we multiply the output of the CBR block by a learnable factor $\beta$ and add it to the current features to get the edges-attentive features Fea:
 \begin{equation}
Fee = Fc \bigotimes U(Fh)  
\end{equation}\begin{equation} Fea = Fc + \beta  CBR(Fee)
\end{equation}where U denotes bilinear upsampling operation, and the $\bigotimes $ represents the point-to-point multiplication. \\
\indent However, the above method presupposes that the higher-level prediction map can accurately locate the target. Therefore, we design a completely symmetric branch, but change the higher-level prediction to higher-level features which are rich in semantic information (e.g. location information), to continuously improve the accuracy of target localization, that is, to continuously optimize the target body. Specifically, Multiply the upsampled higher-level features by current-level features to get body-refined features Fbr. Then, feed Fbr to a CBR block, and multiply the output of the CBR by a learnable $\gamma $, then add the current-level features to get the body-attentive features Fba. Finally, we concatenates the Fea and Fba, then feed the concatenated features into a CBR block to get the refined features Fr. The Fr will be projected to a 1-channel alpha prediction which will  be used as the higher-level prediction for the next SOBE block. \\ \indent The reason we do not use the higher-level prediction map directly to optimize the target body is that, on the one hand, the higher-level prediction map is not necessarily accurate; on the other hand, if we do so, the final output of the model will be extremely dependent on the prediction of the LA-Position module, resulting in poor model robustness. We validated the effectiveness of the SOBE block on three camouflaged object segmentation datasets.

\emph{Output block (Out-blk)} does not use SOBE block because we find that directly feeding the original image as the current-level features into the SOBE block will introduce additional noise. This block first concatenates the original image and the features extracted by the last SOBE block, and then refines the concatenated features by two CBR blocks, and finally outputs a 1-channel alpha prediction and a 3-channel foreground prediction through an Alpha convolution layer and a foreground convolution layer, respectively. The detailed structure is shown on the right  in Fig. 2.
}
 
\subsubsection{Deep Guided Filter}
we adopt a optional Deep Guided Filter (DGF) \cite{2018Fast} module derive from RVM \cite{2021Robust} for high-resolution image matting. When processing high-resolution videos, we first downsample the input before it can be fed into the network for processing. Then, the outputs of the Out-blk, including the low-resolution alpha prediction, foreground prediction, and features for upsampling, as well as the high-resolution input will be fed into the DGF module to generate the high-resolution alpha prediction and foreground prediction,  Note that this module is not required when inputting low-resolution images.

\begin{figure}
\centerline{\includegraphics[width=\columnwidth]{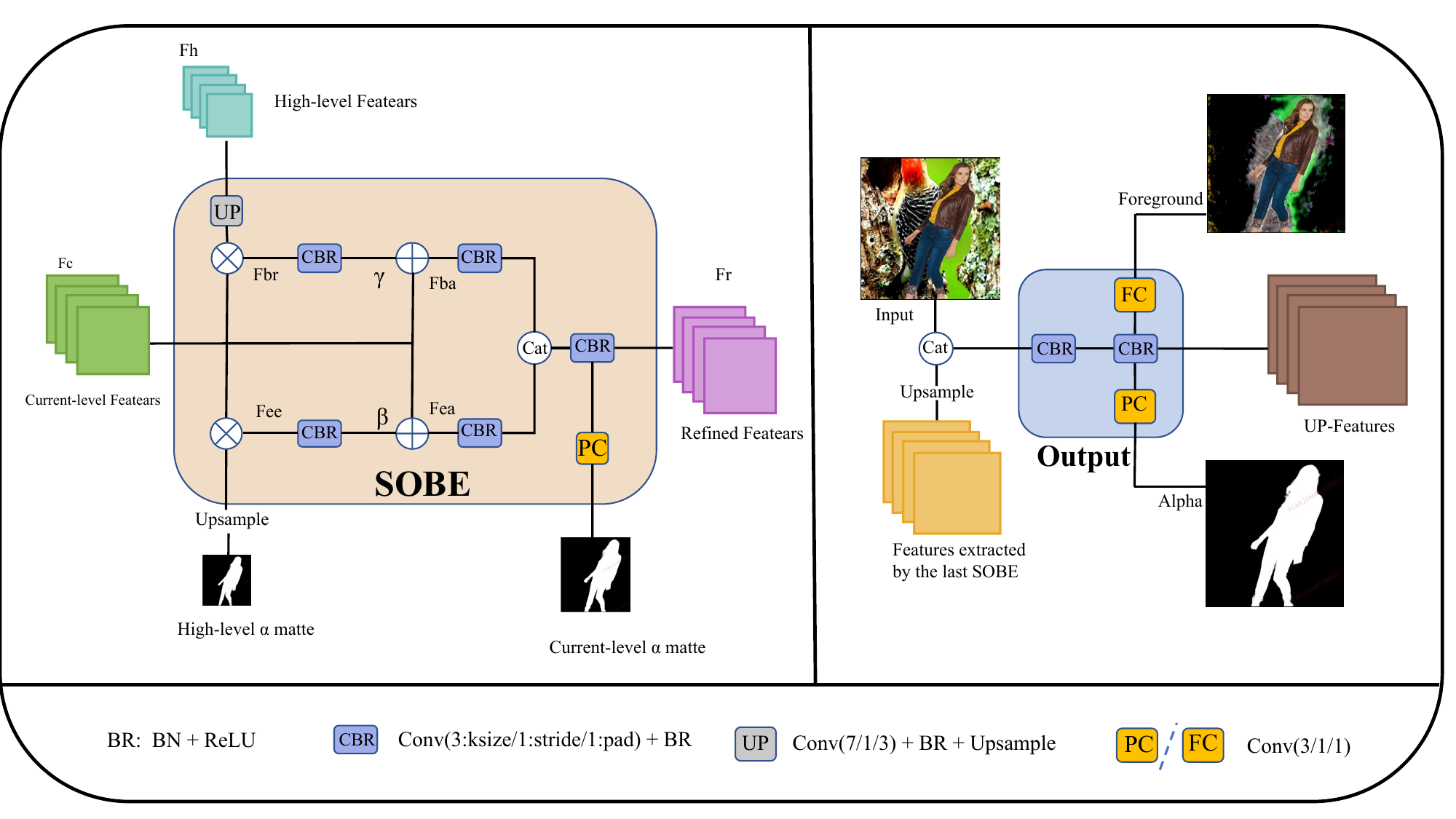}}
\caption{The left is the SOBE block, and the right is the output block(Out-blk). Both PC block and FC block  are convolutional layer and have the same parameters except for the different number of output channels. And the UP-Features is one of the inputs to DGF for generating high-resolution output}
\end{figure} 
\subsection{training strategy}
We train our POBEVM model on VideoMatte240K (VM) \cite{2020Background},  Adobe Image Matting (AIM) \cite{2017Deep} and Distinctions-646 (D646) \cite{2020Attention} datasets. The VM dataset contains low-resolution VideoMatte240K-SD (VM-SD) and high-resolution VideoMatte240K-HD (VM-HD) datasets. Following RVM \cite{2021Robust}, we divided VideoMatte240K into 475/4/5 video clips for training, validation, and testing, and merged D646 and AIM into the Imagematte (IM) dataset, dividing IM into training, validation and testing sets according to its official way. The IM dataset is also processed as low-resolution IM (IM-LR) dataset and high-resolution IM (IM-HR) dataset according to RVM \cite{2021Robust}. The background datasets we use include the DVM Background Set  \cite{2021Deep} processed by \cite{2021Robust}, the Background Dataset - 20k (BG-20k) \cite{2022Bridging}, and 6000 pictures we crawled from the Internet.These background data sets are divided into training, validation and test sets in the ratio of 8:1:1. Of course we also made some appropriate data augmentations, such as affine translation, scale, rotation, brightness, saturation, contrast, hue, etc.

There are five alpha predictions and one foreground prediction. For each alpha prediction, we use the same loss functions, including L1 loss $\mathcal{L}_{l1}^\alpha$, pyramid Laplacian loss $\mathcal{L}_{lap}^\alpha$ \cite{2020Context}, temporal coherence loss  $\mathcal{L}_{ltc}^\alpha$ \cite{2021Deep}. So the loss function of each alpha prediction can be expressed as $\mathcal{L^ \alpha}$ = $\mathcal{L}_{l1}^\alpha$ + $\mathcal{L}_{lap}^\alpha$ + $\mathcal{L}_{ltc}^\alpha$. But for the last alpha prediction, we add an additional loss function, namely our proposed edge-L1 loss (ELL) $\mathcal{L}_{ell}^\alpha$, which first obtains the edge maps of the alpha prediction and corresponding label separately, and then calculates the L1 loss between them. And for the foreground prediction, we use the L1 loss $\mathcal{L}_{l1}^F$ and temporal coherence loss  $\mathcal{L}_{ltc}^F$ on piexls with alpha prediction value greater than 0 following the approach of \cite{2020Background}. So, the loss function of the foreground prediction is $\mathcal{L}^F$ = $\mathcal{L}_{l1}^F$  +   $\mathcal{L}_{ltc}^F$. And our training process can be divided into 4 stages. So, the loss function for each stage can be expressed as: 
\begin{equation}
\mathcal{L}_1 = \mathcal{L}_1^ \alpha + \sum_{i=2}^5 2^{i-2} \mathcal{L}_i^ \alpha + 8 \mathcal{L}^F
\end{equation}
\begin{equation}
\mathcal{L}_{234} = \mathcal{L}_1^ \alpha + \sum_{i=2}^5 2^{i - 2} \mathcal{L}_i^ \alpha + 8 \mathcal{L}^F + 32 \mathcal{L}_{ell}^\alpha
\end{equation}Where $\mathcal{L}_1$ denotes the first training stage, $\mathcal{L}_{234}$ denotes the three other training stages, and $ \mathcal{L}_i^ \alpha $ denotes the \emph{i-th} alpha prediction. For the first stage, we only train our POBEVM on the VM-LR dataset for 15 epochs and change the network learning rate at the 6th epoch. In the second stage, we add the IM-LR dataset to the training for 5 more epochs. And in the third stage, we replace the IM-LR dataset with the VM-HR dataset for another 2 epoch. Finally, we train our model on the VM-LR, IM-LR and IM-HR datasets for 4 epochs.The parameter settings for each training stage are shown in Table \ref{table: training setting}.

\begin{table}[!ht]
    \caption{Training parameter settings at each stage}
    \label{table: training setting}
   \centering
   \scalebox{0.75}
   {\begin{tabular}{|c|c|c|c|c|c|c|c|}
    \hline
   \multirow{2} {*} {Satge }& \multirow{2} {*} {Datasets} & \multirow{2} {*} {Epoch} &\multirow{2} {*} {Batch} & Encoder & Decoder & DGF \\
   \cline{5-7}
  & & & & \multicolumn{3}{|c|}{Learning Rate}   \\
    \hline
    1 & VM-SD & 0-5/6-14& 10 & 1 $e^{-4} $/5  $e^{-5} $ &2 $e^{-4} $ &  \\
     \hline
      2 & VM-SD / IM-LR  & 15-19& 20/20 & 2 $e^{-5} $ & 1 $e^{-4} $ &  \\
     \hline
      3 & VM-SD/ VM-HD & 20-21&20/6 & 1 $e^{-5} $ &   2 $e^{-5}  $&  \multirow{2} {*} {2 $e^{-4} $}\\
    \cline{1-6}
      4 & VM-SD/ IM-(LR/HR) & 22-25&20/6 & 1$e^{-5} $&   1 $e^{-5} $ &  \\
    \hline
    \end{tabular} }
\end{table}

\section{EXPERIMENTS}
In this section, we verify the effectiveness of our proposed SOBE block as well as the POBEVM network through a large number of experiments. The experiments are as follows.

\subsection{Matting experiments}
First, for the target edge, we compared our proposed POBEVM method against state-of-the-art background-based methods BGMv2 \cite{2020Background}, trimap-free method RVM \cite{2021Robust}, and the latest transformer-based method VMFormer \cite{Li2022VMFormerEV}  on three benchmark datasets according to mean absolute difference(MAD), mean square error(MSE), spatial gradient (Grad) \cite{2009A}, connectivity (Conn) \cite{2009A}, and dtSSD \cite{2015Perceptually}. Note that since the dataset we use is included in the training process of RVM and BGMv2, we directly choose their officially given pre-trained models for comparison. For VMFormer, we try to retrain it on our datasets but get worse results, so VMFormer uses its official weights. And The experimental results are shown in Table \ref{table: experimental results for the target edge}. 
\begin{table}[!ht]
    \caption{The experimental results  for the target edge}
    \label{table: experimental results for the target edge}
   \centering
   \scalebox{0.9}
   {
    \begin{tabular}{|c|c|c|c|c|c|c|}
    \hline
  \multirow{2} {*} {Dataset }& \multirow{2} {*} {Method} & \multicolumn{5}{|c|}{Alpha-Edge}\\
\cline{3-7}
& & MAD &MSE &Grad &Conn & dtSSD  \\
    \hline
   \multirow{6} {*}  {VM}&VMFormer & 2.068 & 0.689 &0.537 &0.301  & 1.677\\
   \cline{2-7}
   &BGMv2 & 2.543 & 1.166 &1.061 &0.377 & 1.876\\
  \cline{2-7}
    &RVM & 1.835 & 0.560 &0.484 &0.269 &$ \textbf{1.377} $\\
 \cline{2-7}
    & POBEVM$^{*}  $&1.831 &0.579 &0.510 &0.266& 1.438 \\
\cline{2-7}
    &POBEVM& $ \textbf{1.765} $&$  \textbf{0.548}$&$ \textbf{0.476} $& $\textbf{0.253}$& 1.416 \\
  \hline
     \multirow{6} {*}  {D646}&VMFormer & 5.143 & 1.389 &1.661 &1.315  & 3.613\\
   \cline{2-7}
   &BGMv2 & 4.743 &1.303 &1.506 &1.185 & 3.909 \\
  \cline{2-7}
    &RVM & 4.821&1.258&1.433 &1.237 &$ \textbf{3.350} $\\
 \cline{2-7}
    &POBEVM$^{*} $ & 4.748 & 1.203 & 1.316 & 1.226 & 3.597 \\
\cline{2-7}
    &POBEVM&$ \textbf{4.532} $&$ \textbf{1.126} $&$\textbf{1.220}$&$ \textbf{1.163}$& 3.463 \\
  \hline
      \multirow{6} {*}  {AIM}&VMFormer & 8.755& 3.331 &4.554 &2.305 & 3.923 \\
   \cline{2-7}
   &BGMv2&$ \textbf{6.905} $&$ \textbf{2.186}$ &$\textbf{2.540}$&$\textbf{1.757}$&4.415\\
  \cline{2-7}
    &RVM &8.056 & 2.787&3.209&2.121&$\textbf{3.766}$\\
 \cline{2-7}
    &POBEVM$^{*} $&9.037 &3.465 &3.750 &2.403& 5.128 \\
\cline{2-7}
    &POBEVM & 8.677 &3.222 &3.593 &2.312& 4.935\\
  \hline
  \multicolumn{7}{p{230pt}}{The three benchmark datasets include the VideoMatte240K (VM) \cite{2020Background},  Adobe Image Matting (AIM) and \cite{2017Deep} and Distinctions-646 (D646) \cite{2020Attention} datasets. The VM test dataset is derived from RVM with resolution 512x288, while the AIM and D646 test datasets are synthesized by randomly sampling the foreground and background images in the test set with resolution 512x512. And POBEVM$^{*} $ indicates that ELL function is not used.}
    \end{tabular}
   }
\end{table}

And for the overall matting performance, we did a comparison in the same way, and the results are shown in Table \ref{table: three}
\begin{table}[!ht]
    \caption{Experimental results on the overall matting performance}
    \label{table: three}
   \centering
   \scalebox{0.9}
   {
    \begin{tabular}{|c|c|c|c|c|c|c|}
    \hline
  \multirow{2} {*} {Dataset }& \multirow{2} {*} {Method} & \multicolumn{5}{|c|}{Alpha} \\
\cline{3-7}
& & MAD &MSE &Grad &Conn & dtSSD \\
    \hline
   \multirow{6} {*}  {VM}&VMFormer & 6.021 & 1.002 &0.749 &0.366  & 1.697 \\
  \cline{2-7}
   &BGMv2& 7.364 & 2.361 &1.968 &0.595& 1.968\\
 \cline{2-7}
    &RVM & 6.080 & 1.480 &0.876 &0.413 &$ \textbf{1.360} $\\
\cline{2-7}
    &POBEVM$^{*}$& 5.857&$  \textbf{1.123 }$ &0.878 &0.371& 1.556\\
\cline{2-7}
    &POBEVM &$ \textbf{5.834} $ & 1.126 &$\textbf{0.856} $&$\textbf{0.361} $& 1.547  \\
  \hline
     \multirow{6} {*}  {D646}  &VMFormer & 13.003& 6.857 &2.797 & 3.105  & 5.390\\
  \cline{2-7}
   &BGMv2 & 6.243 & 2.142 &3.163 &1.515 & 4.811 \\
\cline{2-7}
    &RVM & 5.944 & 1.772&2.299&1.459 &$ \textbf{3.768}$\\
\cline{2-7}
    &POBEVM$^{*} $ & 6.043 &1.730 &2.293 &1.464& 4.263\\
\cline{2-7}
    &POBEVM& $ \textbf{5.812} $&$ \textbf{1.624} $&$\textbf{2.149}$ &$ \textbf{1.373} $& 4.143 \\
  \hline
      \multirow{6} {*}  {AIM}&VMFormer & 26.708 & 16.510 &6.426 &6.571  & 6.775 \\
\cline{2-7}
   &BGMv2 &$\textbf{8.572} $&$ \textbf{2.995} $&$\textbf{3.902}$ &$\textbf{2.117} $& 5.189\\
\cline{2-7}
    &RVM & 11.548& 4.901&4.235&2.964&$ \textbf{4.362} $\\
\cline{2-7}
    &POBEVM$^{*} $&13.902 &6.182 &4.808 &3.581& 6.731\\
\cline{2-7}
    &POBEVM& 14.119 &6.195 &4.703 &3.584& 6.603 \\
  \hline
    \end{tabular}
}  
\end{table}

The above two experimental results show that our POBEVM network can achieve state-of-the-art performance compared to prior trimap-free matting methods on both D646 and VM dataset. And our ELL loss function can effectively improve the optimization of the model for the target edge.\\ \indent We also visualize some composited images from the test set and the corresponding alpha matte predictions as well as the edge alpha matte predictions as shown in Fig.3.
\begin{figure}
\label{visual}
\centerline{\includegraphics[width=\columnwidth]{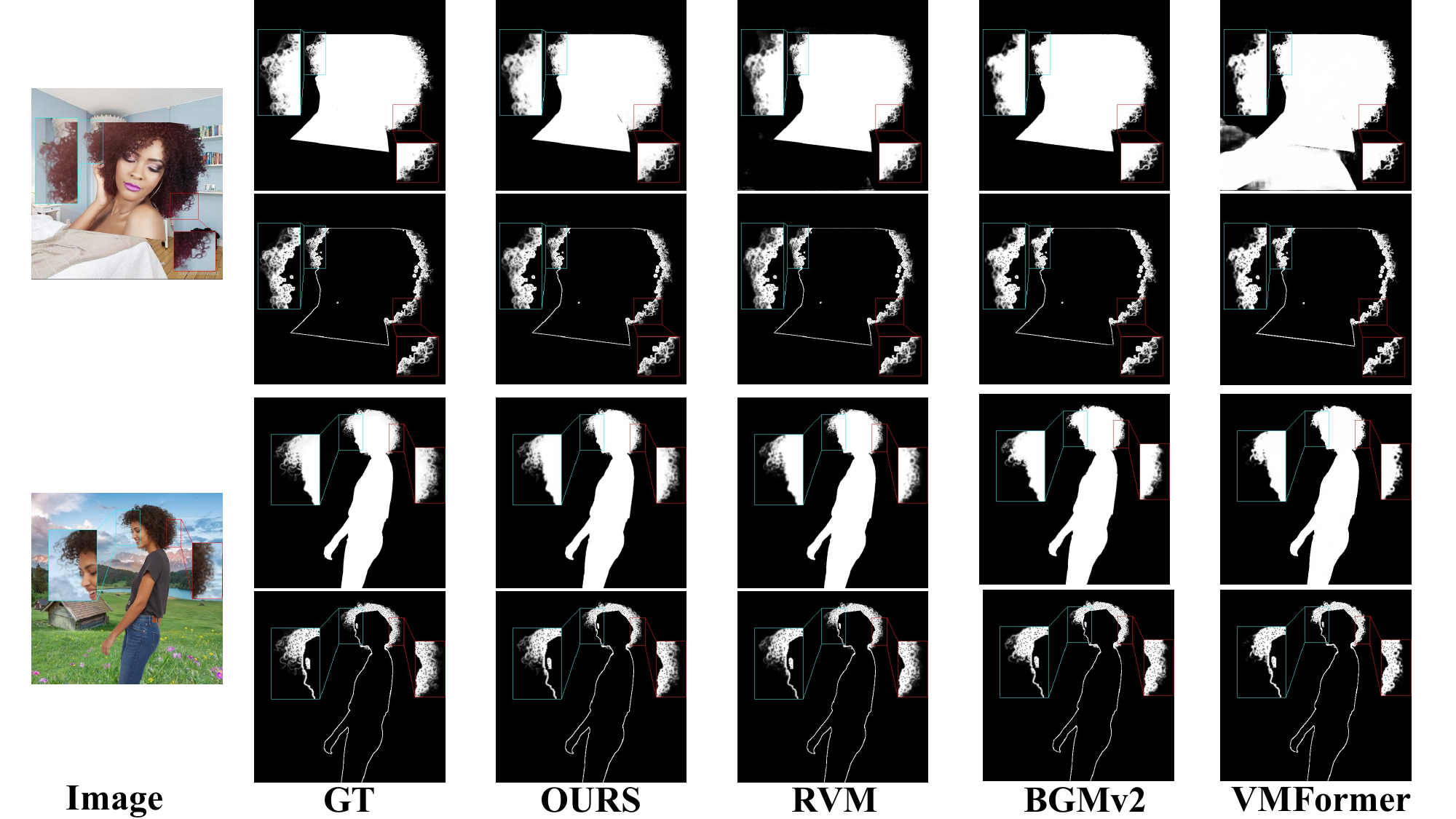}}
\caption{Visualization of alpha matte predictions from BGMv2, RVM, VMFormer and POBEVM(OURS). Our method produces more detailed alpha compared to others.}
\end{figure} 
\subsection{Segmentation experiments}
To verify the effectiveness of the SOBE block, we replaced the Focus block of the PFNet \cite{2021Camouflaged} which is used for camouflaged object segmentation with our SOBE block and trained on the same datasets using the PFNet training strategy, only modify the epoch from 45 to 100. And compared our method with other 4 state-of-the-art methods in the relevant fields in terms of the structure measure $\emph{S}_{\alpha}$  (larger is better), the adaptive E-measure $\emph{E}_{\phi}^{ad}$ (larger is better), the weighted F-measure $\emph{F}_{\beta}^{w}$ (larger is better), and the mean absolute error $\emph{M}$ (smaller is better) on three benchmark
datasets: CHAMELEON \cite{CHM}, CAMO \cite{2019Anabranch}, and COD10K
\cite{2020Camouflaged}. CHAMELEON has 76 images collected from the Internet. CAMO contains 250 testing images. And COD10K contains 2,026 testing images.
\begin{table}[!ht]
    \caption{Experimental results on the camouflaged object segmentation}
    \label{table: four}
   \centering
   \scalebox{0.77}
   {
    \begin{tabular}{|c|c|c|c|c|c|}
    \hline
  \multirow{2} {*} {Dataset }& \multirow{2} {*} {Method} & \multicolumn{4}{|c|}{Segmentation}\\
\cline{3-6}
& &  $\emph{S}_{\alpha} \uparrow $ & $\emph{E}_{\phi}^{ad} \uparrow$ & $\emph{F}_{\beta}^{w} \uparrow $& $\emph{M} \downarrow$\\
    \hline
   \multirow{5} {*}  {CHAMELEON(76 images)}&DSC(2018)& 0.850 &0.888 &0.714&0.050\\
   \cline{2-6}
   &PFANet(2019) & 0.679 & 0.732 & 0.378 & 0.144 \\
   \cline{2-6}
   &SINet(2020) & 0.869 & 0.899 &0.740&0.044  \\
  \cline{2-6}
    &PFNet(2021) & 0.882 & 0.942 &0.81&0.033  \\
     \cline{2-6}
    &PFNet+SOBE(OURS)  &$ \textbf{0.901} $&  $\textbf{0.951} $&$ \textbf{0.839}$  &$ \textbf{0.029}$  \\
  \hline
     \multirow{5} {*}  {CAMO-Test (250 images)}&DSC(2018) &0.736 & 0.830 &0.592 &0.105  \\
   \cline{2-6}
   &PFANet(2019) & 0.659 & 0.735 & 0.391 & 0.172 \\
   \cline{2-6}
   &SINet(2020) &0.751 & 0.834 &0.606&0.100  \\
  \cline{2-6}
    &PFNet(2021) & $ \textbf{0.782} $& $ \textbf{0.852 }$&$\textbf{0.695}$& $\textbf{0.085}$   \\
     \cline{2-6}
    &PFNet+SOBE(OURS)  &0.780 & $ \textbf{0.852 }$& 0.677& 0.087\\
  \hline
      \multirow{6} {*}  {COD10K-Test (2,026 images)}&DSC(2018) &0.758 & 0.788 &0.542 &0.052  \\
   \cline{2-6}
   &PFANet(2019) &0.636 & 0.619 & 0.286 & 0.128 \\
   \cline{2-6}
   &SINet(2020) &0.771 & 0.797 &0.551&0.051  \\
  \cline{2-6}
    &PFNet(2021) &  0.800 &0.868 &0.660& 0.040 \\
     \cline{2-6}
    &PFNet+SOBE(OURS) & $\textbf{0.807 }$&  $\textbf{0.874}$& $ \textbf{0.665}$&$ \textbf{0.039}$\\
  \hline
    \end{tabular}
    }
\end{table}

\section{CONCLUSION}
In this paper, we propose a module SOBE and network POBEVM based on CNNs for video matting, and also propose a loss function ELL. They are all specifically designed to optimize the target edges. Extensive experiments have proved the rationality and effectiveness of our proposed methods. We also note that the POBEVM network does not perform well on the AIM dataset. and our preliminary analysis suggests that this is due to the uneven distribution of the dataset during the training process, and we will conduct further experiments to investigate this problem.

\bibliographystyle{IEEEtran}
\bibliography{ref}   

\end{document}